\documentclass{article}

\usepackage{arxiv1}

\usepackage[utf8]{inputenc} 
\usepackage[T1]{fontenc}    
\usepackage{hyperref}       
\usepackage{url}            
\usepackage{booktabs}       
\usepackage{amsfonts}       
\usepackage{nicefrac}       
\usepackage{microtype}      
\usepackage{lipsum}
\usepackage{multirow}
\usepackage{graphicx}
\usepackage[font=footnotesize]{subcaption}
\usepackage{cite}

\newcommand\blfootnote[1]{%
  \begingroup
  \renewcommand\thefootnote{}\footnote{#1}%
  \addtocounter{footnote}{-1}%
  \endgroup
}

\title{SFU-Store-Nav: A Multimodal Dataset for Indoor Human Navigation}

\author{
  Zhitian Zhang \\
  School of Computing Science\\
  Simon Fraser University\\
  Burnaby, BC Canada \\
  \texttt{zhitianz@sfu.ca} \\
   \And
  Jimin Rhim \\
  School of Computing Science\\
  Simon Fraser University\\
  Burnaby, BC Canada \\
  \texttt{jrhim@sfu.ca} \\
   \And
  Taher Ahmadi \\
  School of Computing Science\\
  Simon Fraser University\\
  Burnaby, BC Canada \\
  \texttt{tahera@sfu.ca} \\
   \And
  Kefan Yang \\
  School of Computing Science\\
  Simon Fraser University\\
  Burnaby, BC Canada \\
  \texttt{kefany@sfu.ca} \\
  \And
  Angelica Lim \\
  School of Computing Science\\
  Simon Fraser University\\
  Burnaby, BC Canada \\
  \texttt{angelica@sfu.ca} \\
  \And
  Mo Chen \\
  School of Computing Science\\
  Simon Fraser University\\
  Burnaby, BC Canada \\
  \texttt{mochen@sfu.ca} \\
}

\begin{document}
\maketitle

\begin{abstract}
This article describes a dataset collected in a set of experiments that involves human participants and a
robot. The set of experiments was conducted in the computing science robotics lab in Simon Fraser
University, Burnaby, BC, Canada, and its aim is to gather data containing common gestures, movements,
and other behaviours that may indicate humans’ navigational intent relevant for autonomous robot
navigation. The experiment simulates a shopping scenario where human participants come in to pick up
items from his/her shopping list and interact with a Pepper robot that is programmed to help the human
participant. We collected visual data and motion capture data from 108 human participants. The visual data
contains live recordings of the experiments and the motion capture data contains the position and orientation
of the human participants in world coordinates. This dataset could be valuable for researchers in the
robotics, machine learning and computer vision community.
\end{abstract}

\blfootnote{This paper is submitted to Data In Brief Journal}
\begin{table}[h!]
 \caption{\textbf{Specification Table}}
  \centering
  \begin{tabular}{p{0.3\textwidth} | p{0.6\textwidth}}
   \hline
    Subject    & Computer Vision and Pattern Recognition   \\
    \hline
    Specific subject area  & Human Intent Inference      \\
    \hline
    Type of data    & Video data, in AVI file format. \\
    & Vicon motion tracking data, in CSV file format.     \\
    \hline
    How data were acquired     & Hardware:\\
    &Logitech Webcams\\
    &Vicon Motion Capture System\\
    &Softbank Pepper robot\\
    &Software:\\
    &Robot Operating System \cite{quigley2009ros}\\
    &Python      \\
    \hline
    Data format & Extracted\\
    &Processed\\
    \hline
    Parameters for data collection & The data collection experiments were conducted under two conditions: one participant and two-participant. For one-participant trial, a single participant performs the experimental task by himself/herself. And for the two participant trial, each of the two participants perform the experimental tasks in the physical presence of the other. \\
    \hline
    Data source location & Institution: Simon Fraser University\\
    &City/Town/Region: Burnaby, British Columbia\\
    &Country: Canada\\
    \hline
    Data accessibility & Repository name: SFU Vault\\
    &URL to data: https://vault.sfu.ca/index.php/s/QCSRW2sXlGicHIb\\
    &Instructions for accessing these data:\\
    &URL to the data is password protected. Access to the data can be obtained from https://www.rosielab.ca/datasets/sfu-store-nav. Users will need to fill out a form to obtain the password. More detailed instruction can be found on the website. \\
    \hline
    Related research article & Author’s name: Zhitian Zhang\\
    &Title: Towards a Multimodal and Context-Aware Framework for Human Navigational Intent Inference\\
    &Conference: The 22nd ACM International Conference on Multimodal Interaction (ICMI2020)\\
    &DOI: 10.1145/3382507.3421156\\
    \hline

  \end{tabular}
  \label{tab:table}
\end{table}

\section{Value of the Data}
\begin{itemize}
    \item Current human-robot interaction datasets are very limited in terms of data modality. Our dataset contains not only the visual information collected from the experiments, but also important information such as participants’ positions and head orientations. These data provide great value for the robotics research community.
    \item Researchers in computer vision, machine learning, affective computing and robotics can all benefit from this dataset. 
    \item In robotics research, this data can be used to infer human navigational intent and help create better and safer robot planning algorithms. 
    \item Computer vision researchers can use the dataset for topics like trajectory prediction, human body pose estimation, etc and benefit from our multi-camera setup.
    \item Compared to other currently available human movement dataset, our dataset captured human’s ground truth location in real-world coordinates, in addition to providing visual information.
    \item Our data from two-participant trials may also capture possible useful information on human-human interactions when a robot is present. 
    \item In our data collection experiment, we purposely designed a scenario where the human participants will become confused and ask robot for help. Such scenarios, with a realistic shopping environment involving confused humans and robot that are ready to assist humans, are valuable to study. 
    
\end{itemize}

\section{Data Description}
\subsection{Raw Data}
Raw data is captured through the Robot Operating System (ROS) and stored in ROS bag files. Raw data can be extracted into visual data, human position data and human head orientation data. 
\subsection{Visual Data}
\begin{figure}
    \centering
    \begin{subfigure}{0.4\textwidth}
     \includegraphics[width=\textwidth]{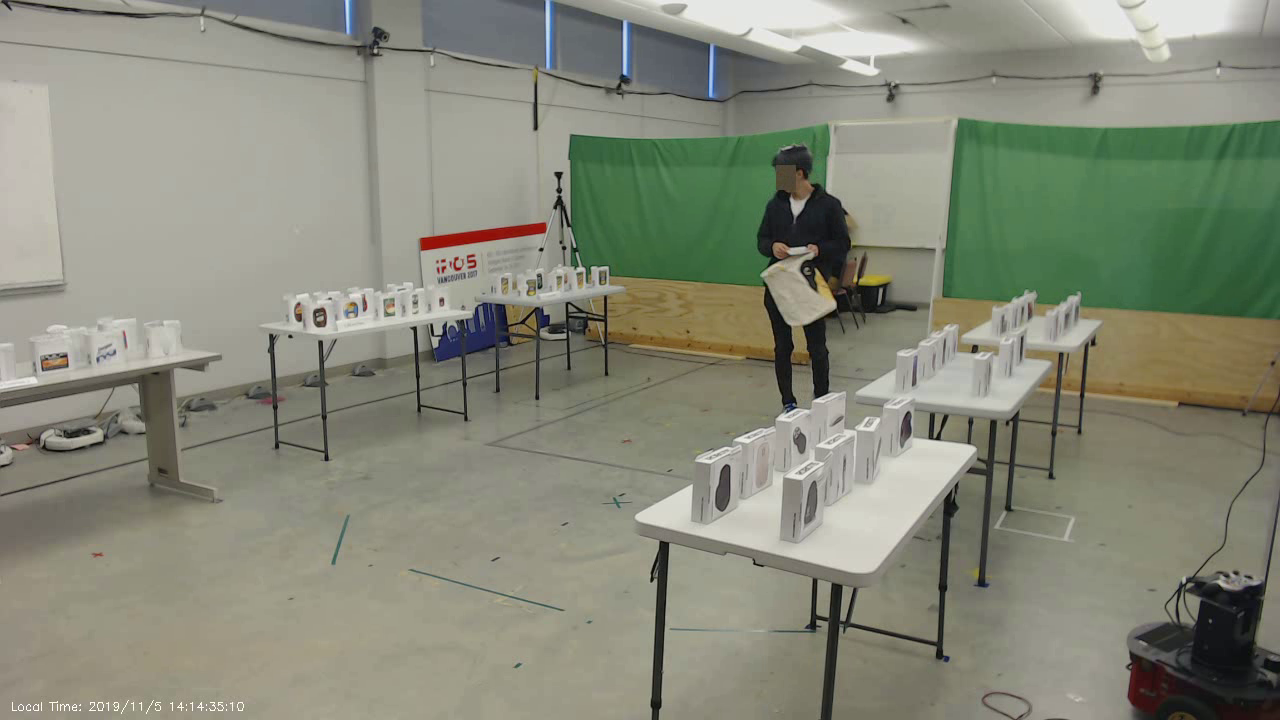}
     \caption{}
    \end{subfigure}
    \begin{subfigure}{0.4\linewidth}
     \includegraphics[width=\linewidth]{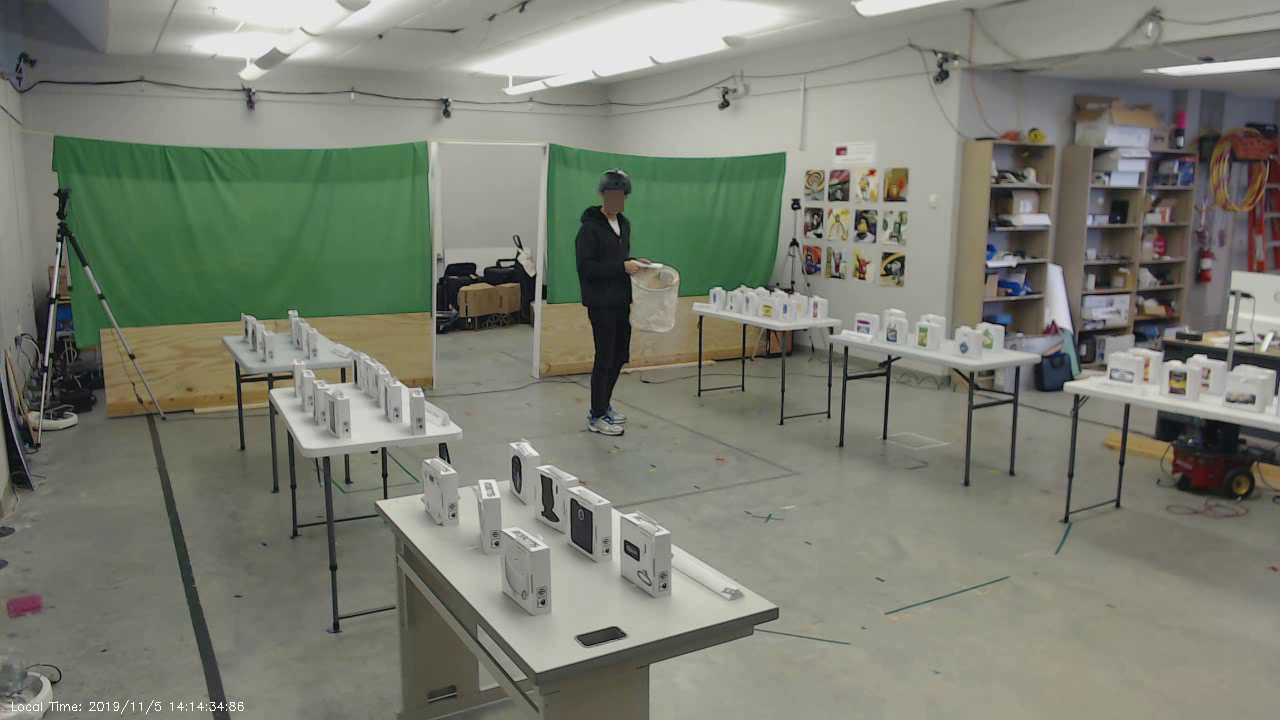}
     \caption{}
    \end{subfigure}
    \begin{subfigure}{0.4\linewidth}
     \includegraphics[width=\linewidth]{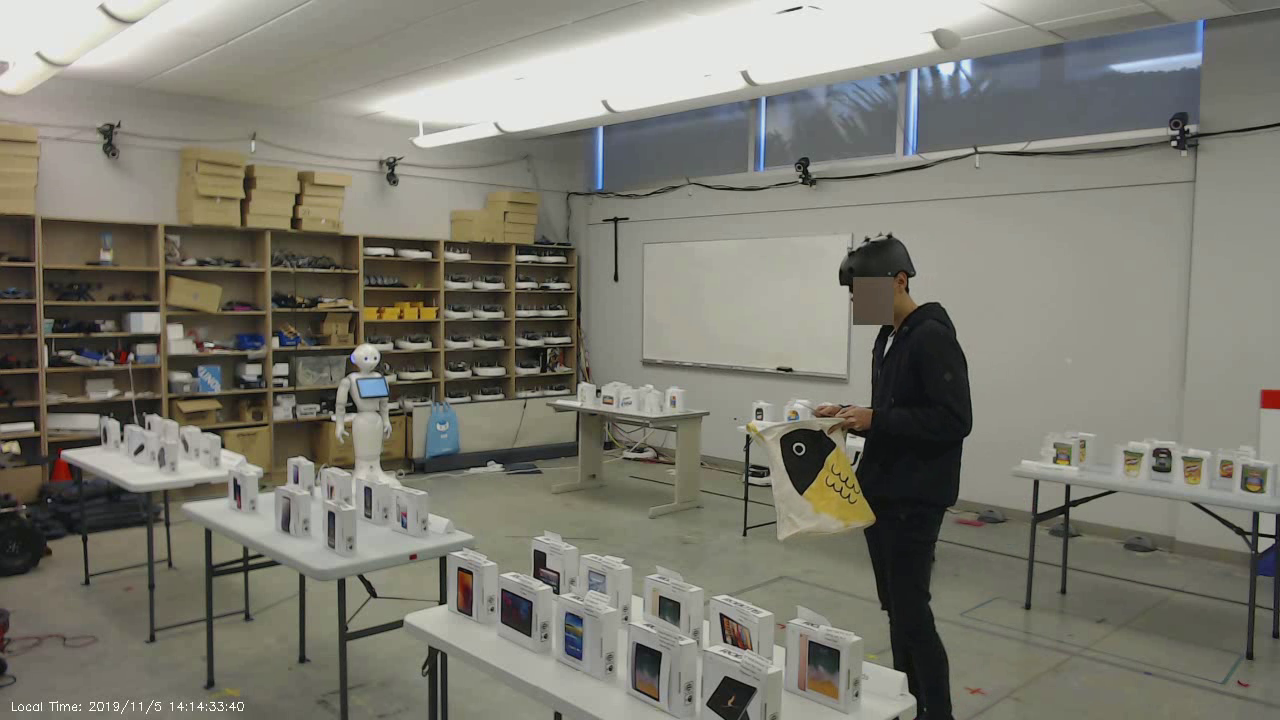}
     \caption{}
    \end{subfigure}
    \begin{subfigure}{0.4\linewidth}
     \includegraphics[width=\linewidth]{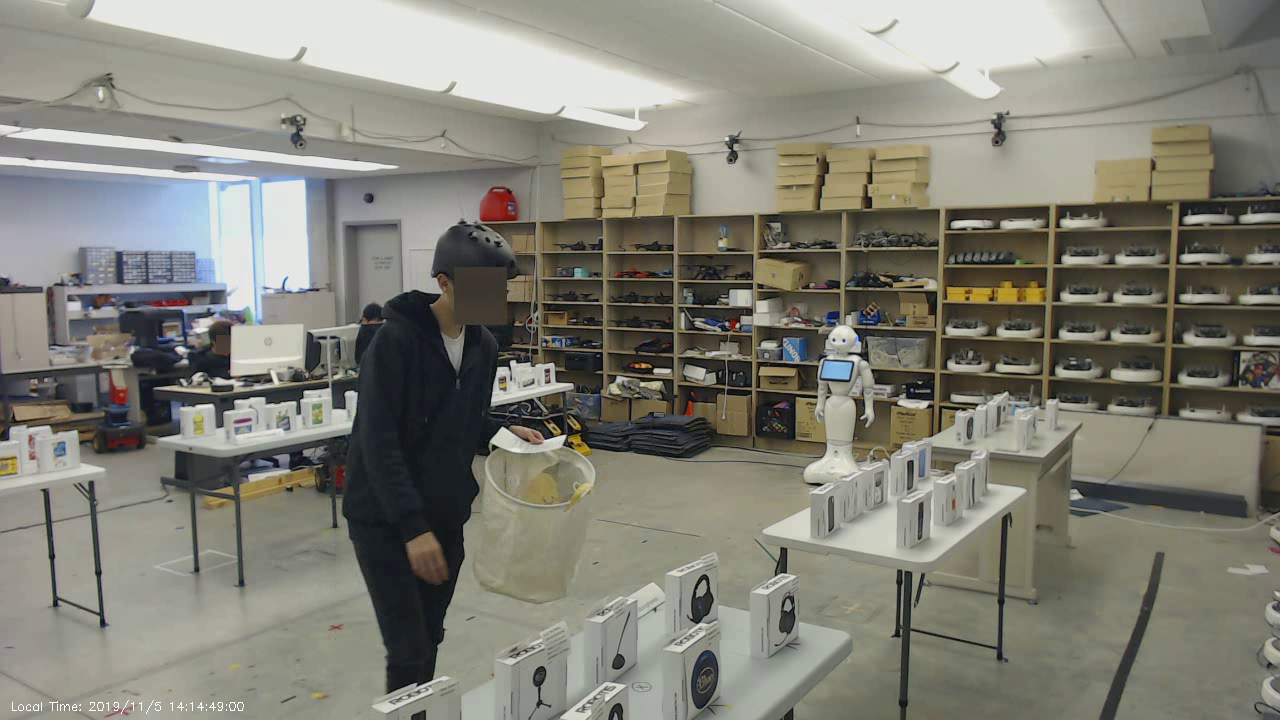}
     \caption{}
    \end{subfigure}
    \caption{\textbf{Anonymized images captured from four webcams.}}
    \label{fig:1}
\end{figure}

\begin{figure}
  \centering
  \includegraphics[width = 0.4\textwidth]{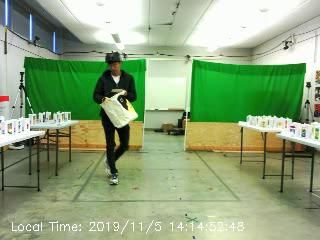}
  \caption{\textbf{Image captured from the Pepper robot.}}
  \label{fig:2}
\end{figure}

There are two sources of visual data: webcams and Pepper robot’s built-in camera. We record the experiment with four webcams and Pepper robot’s camera. Figure \ref{fig:1} shows the images that are captured through webcams at the same moment. Figure \ref{fig:2} is an image captured by the Pepper robot. Images are extracted from the ROS bag files and compressed into video files using ROS and OpenCV \cite{bradski2008learning}. Every frame in the video are labelled with the time, and we also provide a CSV file that records total frame numbers and their corresponding time stamps for each video. The resolution of the video captured from webcams is 1280 x 720 pixels. And the resolution for videos captured by robot is 320 x 240 pixels. To protect the privacy of the participants that involves in the experiment, we used an open source tool \cite{li2019dsfd} to anonymize videos captured from webcams as shown in Figure 1. Videos captured from the Pepper robot are not anonymized due to low resolution.

\begin{figure}
  \centering
  \includegraphics[width = 0.4\textwidth]{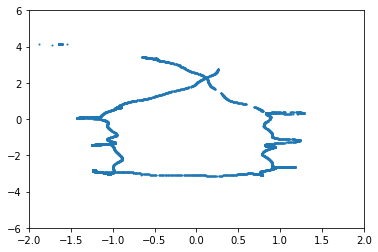}
  \caption{\textbf{Human position data captured from Vicon system.}}
  \label{fig:3}
\end{figure}

\subsection{Human position and head orientation data}
Participants’ real time position data and head orientation data are obtained through a Vicon motion capture system using the ROS framework. Both position and head orientation are in real world coordinates and stored in CSV files. Each position/head orientation data point corresponds to a set of images with the same time stamp. That means our position/head orientation data are synchronized with the image data in time. Position data is described in (X, Y) format and head orientation is described in (roll, pitch, yaw) format. Figure 3 shows the position data for a whole experiment trial. 

\section{Experimental Design, Materials and Methods}
The aim of the experiment is to gather data containing common gestures that may indicate humans’ navigational intent relevant for autonomous robotic navigation. The experiment simulated shopping in an electronic store and a grocery store, an activity that encourages natural human movements and provides an opportunity to interact with a robot.  In real life, people do not always shop alone. Moreover, different shopping experiences (e.g., shopping alone, shopping in a familiar or unfamiliar environment) would lead to different navigation patterns. Therefore, we created two study conditions: one-participant trial and two-participants trail. Details of the study conditions are described below: 
\begin{itemize}
    \item One-participant trial: 
    \begin{itemize}
        \item A single participant performs the study tasks in the presence of a robot, for the first time, as a “newcomer” of an unfamiliar environment.
        \item This participant will repeat the study tasks, with a different shopping list, as an “oldcomer” who is familiar with the store environment. 
    \end{itemize}
    \item Two-participants trial: Two participants conduct study tasks in the physical presence of another. One participant is a newcomer conducting the task for the first time, and the other participant will be an oldcomer who has conducted the task before. They will also conduct the study task in the presence of a robot.
\end{itemize}

\subsection{Participants}
The study was conducted in Simon Fraser University, Burnaby, BC, Canada. There were 108 participants total, and each received a \$10 CAD gift card. Participants were recruited via class announcements and school mailing lists. There are 36 one-participant trials and 36 two-participants trial in total. Although the number of trials is the same, the number of participants involved in the one-participant trials is significantly higher than those involved in two-participant trials.

\subsection{Materials and Experimental Set up }
We used a Pepper robot by Softbank robotics, two computers, and a Vicon motion capture system in this experiment. Pepper was used because it is a programmable robot which is designed to interact with human users, which allows the data to be representative of how humans may behave while interacting with a robot. The robot was programmed using NAOqi SDK \footnote{http://doc.aldebaran.com/2-4/dev/python/index.html}, a Python based development tool for the Pepper. One computer was used by the Pepper operator to coordinate Pepper’s dialogues, which was pre-programmed to give the illusion that the robot is autonomously responding to the participants. The other computer was used to record the frequency and duration of participants’ interaction with the robot. A total of five cameras recorded the participants’ interactions with Pepper.  Four web cameras were placed at each corner of the experiment room to capture full movements of participants, and the built-in camera of Pepper captured human behaviors from the Pepper's perspective. To gather data for the Vicon motion capture system, we placed reflective markers on a biking helmet which each participant is instructed to wear during the experiments. For two-participant trials, the reflective markers were placed in unique patterns to distinguish two users. A sample experiment setup layout is shown in Figure 4.

\begin{figure}
  \centering
  \includegraphics[width = 0.5\textwidth]{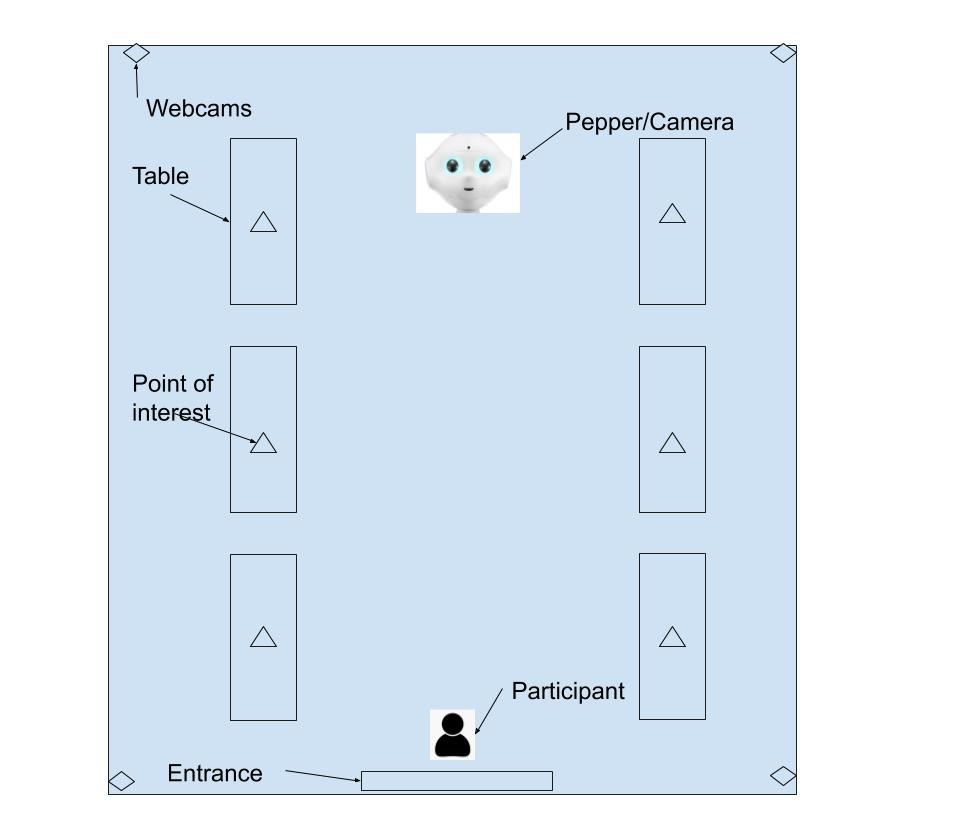}
  \caption{\textbf{Experiment Set up}}
  \label{fig:4}
\end{figure}

\section{Procedure}
Participants signed a consent form before beginning the task. If they agreed to participate in the study, we provided a brief overview of the research and provided study protocol. The participants were provided with 4 to 8 different shopping lists and were asked to find all the items on the provided list. Each list had six items and one or two items were intentionally missing to encourage the participant to interact with the robot. Participants were advised to ask the robot for help if needed when finding items. During the experiments, participants were asked to wear the biking helmet with Vicon markers throughout the task so that their position and head orientation can be measured by the Vicon motion capture system. The study was conducted using a partial Wizard of Oz method. As the robot was not fully autonomous, when participants approached and asked questions to the robot, researcher played pre-recorded scripts of the robot to answer questions. Some examples of the scripts that are used for robots are listed below:
\begin{itemize}
    \item How can I help you today?
    \item That’s a great question, let me check it out for you.
    \item The item you are looking for is on Table A.
    \item Sorry the item you are looking for is out of stock.
\end{itemize}

\section{Ethics Statement}
The study was approved by SFU ethics board (Study number: 2019S027). We confirm that informed consent was obtained for experimentation with human participants.

\section{Acknowledgments}
This work was supported under the Huawei-SFU Joint Lab Project (R569337).

\section{Declaration of Competing Interest}
The authors declare that they have no known competing financial interests or personal relationships which have, or could be perceived to have, influenced the work reported in this article.



\bibliographystyle{unsrt} 

\end{document}